\documentclass[sigconf]{acmart}

\usepackage{adjustbox}
\usepackage{arydshln}
\usepackage{color}
\usepackage{pifont}

\newcommand{\Fmacroavg}{$F1\textnormal{-}Macro_{AVG}$}

\copyrightyear{2024}
\acmYear{2024}
\setcopyright{rightsretained}
\acmConference[CIKM '24]{Proceedings of the 33rd ACM International Conference on Information and Knowledge Management}{October 21--25, 2024}{Boise, ID, USA}
\acmBooktitle{Proceedings of the 33rd ACM International Conference on Information and Knowledge Management (CIKM '24), October 21--25, 2024, Boise, ID, USA}
\acmDOI{10.1145/3627673.3679167}
\acmISBN{979-8-4007-0436-9/24/10}

\makeatletter
\gdef\@copyrightpermission{
  \begin{minipage}{0.3\columnwidth}
    \href{https://creativecommons.org/licenses/by-nd/4.0/}{\includegraphics[width=0.90\textwidth]{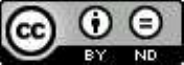}}
  \end{minipage}\hfill
  \begin{minipage}{0.7\columnwidth}
    \href{https://creativecommons.org/licenses/by-nd/4.0/}{This work is licensed under a Creative Commons Attribution-NoDerivs International 4.0 License.}
  \end{minipage}
  \vspace{5pt}
}
\makeatother

\makeatletter
\newcommand\footnoteref[1]{\protected@xdef\@thefnmark{\ref{#1}}\@footnotemark}
\makeatother

\begin{document}

\title{EUvsDisinfo: A Dataset for Multilingual Detection of Pro-Kremlin Disinformation in News Articles}
\author{Jo\~{a}o A. Leite}
\affiliation{%
  \institution{The University of Sheffield}
  \city{Sheffield}
  \country{United Kingdom}}
\email{jaleite1@sheffield.ac.uk}

\author{Olesya Razuvayevskaya}
\affiliation{%
  \institution{The University of Sheffield}
  \city{Sheffield}
  \country{United Kingdom}}
\email{o.razuvayevskaya@sheffield.ac.uk}

\author{Kalina Bontcheva}
\affiliation{%
  \institution{The University of Sheffield}
  \city{Sheffield}
  \country{United Kingdom}}
\email{k.bontcheva@sheffield.ac.uk}

\author{Carolina Scarton}
\affiliation{%
  \institution{The University of Sheffield}
  \city{Sheffield}
  \country{United Kingdom}}
\email{c.scarton@sheffield.ac.uk}

\renewcommand{\shortauthors}{Jo\~{a}o A. Leite, Olesya Razuvayevskaya, Kalina Bontcheva, and Carolina Scarton}
\begin{abstract}
This work introduces EUvsDisinfo, a multilingual dataset of disinformation articles originating from pro-Kremlin outlets, along with trustworthy articles from credible / less biased sources. It is sourced directly from the debunk articles written by experts leading the EUvsDisinfo project. Our dataset is the largest to-date resource in terms of the overall number of articles and distinct languages. It also provides the largest topical and temporal coverage. Using this dataset, we investigate the dissemination of pro-Kremlin disinformation across different languages, uncovering language-specific patterns targeting certain disinformation topics. We further analyse the evolution of topic distribution over an eight-year period, noting a significant surge in disinformation content before the full-scale invasion of Ukraine in 2022. Lastly, we demonstrate the dataset's applicability in training models to effectively distinguish between disinformation and trustworthy content in multilingual settings. 
\end{abstract}

\ccsdesc[500]{Computing methodologies~Natural language processing}
\ccsdesc[500]{Applied computing~Law, social and behavioral sciences}

\keywords{Disinformation, Dataset, pro-Kremlin, Classification, News Articles}


\maketitle

\section{Introduction} \label{sec:intro}

\noindent \textit{Information warfare} is a special case of disinformation
that aims at manipulating public opinion to achieve military or political objectives \cite{thornton2015changing, taylor2013munitions, lanoszka2016russian}. Specifically, pro-Kremlin information warfare has been a topic of study across different disciplines \cite{abrams2016beyond, renz2016russia, pamment2016digital, linvill2019russians}. Since the annexation of Crimea, the spread of pro-Kremlin disinformation in Europe has intensified, encompassing many languages and even influencing search engines \cite{prop2, 10.1093/jogss/ogac004, jarynowski2023biological, williams2023search}.


\begin{table*}[ht]
\centering
\caption{Overview of multilingual article-level content verification datasets.}
\scalebox{0.9}{
\begin{adjustbox}{width=\linewidth}
\begin{tabular}{@{}lclclcc@{}}
\toprule
\textbf{Dataset} & \textbf{Size} & \textbf{Labelling} & \# \textbf{Languages} & \textbf{Multilinguality} & \textbf{Topic(s)} & \textbf{Period} \\ \midrule
\begin{tabular}[c]{@{}l@{}}TALLIP-FakeNews \cite{momchil-et-al-2016-tallip} \end{tabular} & $2,304$ & Manual & $3$ & Translation-based & COVID-19 & 4 months  \\
\begin{tabular}[c]{@{}l@{}}MM-Covid \cite{yichuan-et-al-mmcovid} \end{tabular} & $11,565$ & Fact-checkers + Manual & $6$ & Inherent & COVID-19 & 6 months \\
\begin{tabular}[c]{@{}l@{}}FakeCovid \cite{shahi2006multilingual} \end{tabular} & $5,182$ & Fact-checkers & $40$ & Inherent & COVID-19 & 4 months \\
\begin{tabular}[c]{@{}l@{}}Multiverse \cite{dementieva2023multiverse} \end{tabular} & $3,009$ & Source-based & $5$ & Semi-translation-based & \begin{tabular}{@{}c@{}} Celebrities, Science, \\ Politics, Culture, World \end{tabular} & Unknown \\
\midrule
EUvsDisinfo (ours) & $18,249$ & Fact-checkers & $42$ & Inherent & \begin{tabular}{@{}c@{}}508 topics originating from\\ Pro-Kremlin disinformation\end{tabular} & 8.5 years \\ \bottomrule
\end{tabular}
\end{adjustbox}
}
\label{tab:datasets}
\end{table*}

Combating pro-Kremlin disinformation articles presents significant challenges due to its dissemination across various languages, during an extensive period of time, and covering a wide range of different topics. Current disinformation datasets 
are not suitable for training machine learning models aimed to address this task since they lack the combination of such characteristics. Most datasets focus on analysing short claims \cite{thorne-etal-2018-fever, wang2017liar} or social media posts \cite{potthast-etal-2018-stylometric,zubiaga2016analysing}, whilst the majority of the existing article-level disinformation corpora consist of monolingual data \cite{hossain2020banfakenews, shu2020fakenewsnet, karadzhov-etal-2017-hackfakenews, momchil-et-al-2016-newscredibility, wang-et-al-2020-wefend, golbeck2018fake, perez-rosas-etal-2018-automatic, shu2020fakenewsnet, shahi2021overview}. A significantly smaller amount of multilingual datasets is available \cite{momchil-et-al-2016-tallip, dementieva2023multiverse, yichuan-et-al-mmcovid, shahi2006multilingual}, and they contain the following limitations: (i) representing parallel translated data \cite{momchil-et-al-2016-tallip, dementieva2023multiverse}, (ii) covering short periods of time ($\leq$ 1 year) \cite{momchil-et-al-2016-tallip, yichuan-et-al-mmcovid, shahi2006multilingual}, (iii) containing only few articles (around $1-2K$) \cite{momchil-et-al-2016-tallip}, and (iv) focusing on narrow topics \cite{momchil-et-al-2016-tallip, yichuan-et-al-mmcovid, shahi2006multilingual}. To the best of our knowledge, the work by \citet{solopova-etal-2023-evolution} is the only dataset addressing the theme of pro-Kremlin \textit{information warfare}. Nevertheless, their dataset is targeted towards the detection of a
pro-Western vs pro-Kremlin stance rather than disinformation. 


In this work, we leverage the journalistic investigation carried out by specialists to generate a large and diverse multilingual dataset containing 
disinformation articles from pro-Kremlin outlets and trustworthy counterparts from reliable / less biased sources.
To produce such a dataset, we use the debunk articles written by EUvsDisinfo\footnote{\url{https://euvsdisinfo.eu}}, who since 2015 has been providing an EU-wide response to information warfare by debunking disinformation narratives across EU-member states. Our dataset is the largest ($18,249$ articles) to-date, most topically diverse ($508$ topics 
manually assigned by EUvsDisinfo),
spans over the longest period of time ($8.5$ years), and is the most diverse in terms of languages ($42$ languages) compared to other article-level multilingual disinformation detection datasets (see Table \ref{tab:datasets} for an overview of related datasets in comparison to ours). Our key contributions are: (i) The novel multilingual disinformation dataset\footnote{Dataset: \url{https://doi.org/10.5281/zenodo.10514307}}, becoming the largest and most diverse article-level set in terms of language, topics, and time periods. (ii) The analysis of the prominence of pro-Kremlin disinformation topics across different languages and time periods. (iii) The evaluation of several baseline models using our novel dataset for the task of multilingual binary disinformation detection. Our dataset and code are made publicly available, along with documentation, as supplementary material to facilitate reproducibility\footnote{\label{sup}Supplementary material: \url{https://github.com/JAugusto97/euvsdisinfo}}.

\section{Methodology} \label{sec:methodology}
\begin{figure}[h]
  \begin{adjustbox}{width=0.9\columnwidth}
    \includegraphics{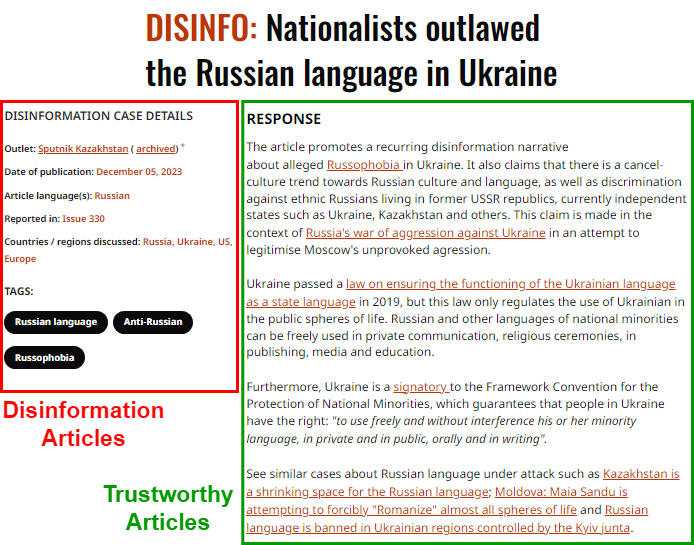}
  \end{adjustbox}
  \caption{Example of debunk article by EUvsDisinfo.}
  \label{fig:euvsdisinfo}
\end{figure}

Our dataset leverages the structured information of the debunking articles published by EUvsDisinfo (see Figure~\ref{fig:euvsdisinfo} for an example). It consists of two classes of articles: disinformation and trustworthy. Disinformation articles are acquired directly from the links mentioned on the left hand-side of the EUvsDisinfo debunk page. 
Trustworthy articles are derived from the links to reliable sources within the response section of the article, in which EUvsDisinfo directly debunks the false narrative. Such URLs can lead to news articles, official documents, statements, books, encyclopedia articles, social media posts, and fact-checking articles, among others. 

To verify that EUvsDisinfo does not reference disinformation articles in the response section, we manually inspected a randomly sampled set of 30 debunk articles, and annotated all the URLs mentioned within their response section. In total, 350 URLs were labelled as either \textit{trustworthy} or \textit{potential disinformation}. To do so, we considered the context in which the URL is referenced in the response section. For example, if the URL is referenced after a sentence such as "see similar cases", it is marked as potential disinformation (for reference, see the last paragraph in the response section of Figure \ref{fig:euvsdisinfo}). We identified that all of the URLs marked as \textit{potential disinformation} actually correspond to other debunk articles by EUvsDisinfo or other fact-checking agencies such as Bellingcat and StopFake, instead of directly links to disinformation.


\subsection{Data Collection}\label{sec:datacoll}
Given the wide variety of websites mentioned in EUvsDisinfo debunks, with varying HTML structure, extracting textual content is not trivial.
For this task, we employ the Diffbot API\footnote{\url{https://www.diffbot.com}}, which is a proprietary tool that uses machine learning to extract the content of a given web page. Although Diffbot is a closed-source service, it is free of charge for academic purposes.
We retrieve articles that are no longer available on the web, by using the digital library Wayback Machine\footnote{\url{https://web.archive.org}}. 
Also, the language of each trustworthy article is inferred using Polyglot\footnote{\url{https://github.com/aboSamoor/polyglot}}, as EUvsDisinfo only specifies the languages of the disinformation articles.

\begin{figure*}[ht]
  \scalebox{0.92}{
    \includegraphics[width=\linewidth]{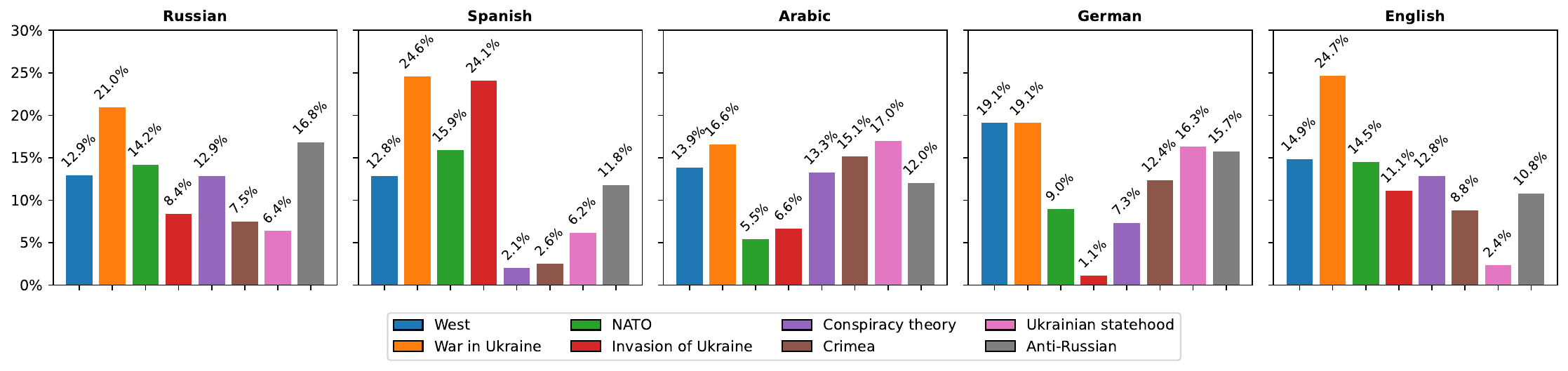}
  }
  \caption{Top 8 most frequent topics across disinformation articles for the top 5 languages.}
  \label{fig:topics_by_language}
\end{figure*}


Upon collecting the contents of all the URLs mentioned on EUvsDisinfo debunks, resulting in $35,839$ articles, we apply three filtering strategies to ensure consistency in the dataset. First, we remove $12,875$ URLs whose domain is not of a news outlet type. This is done to ensure consistency in the theme of disinformation detection of news articles published by news outlets. To do so, we first manually inspect the most frequent domains of the URLs, and label them as one of the following: ``News Outlet'', ``Organisation'', ``Social Media'', ``Fact-checker'', and ``Other''. 
Next, we remove $4,048$ instances referring to error messages, log-in prompts, and paywalls through lexicon-bases rules, and through removing articles with less than $700$ characters, which is significantly shorter than the average article ($6,346$ characters). Lastly, we remove $667$ URLs cited within sentences containing n-grams referring to other fact-checking articles (e.g. "See earlier disinformation cases"). The full list of n-grams can be found in the supplementary materials\footnoteref{sup}.

\subsection{Dataset Overview}
The full dataset has $18,249$ news articles with $10,682$ ($59\%$) and $7,567$ ($41\%$) marked as disinformation and trustworthy, respectively. It contains articles by $2,946$ different publishers on $508$ unique topics, spanning across $8.5$ years, between 06/01/2015 and 01/08/2023. On average, the articles are $6,346$ characters in length and cover $3.8$ different topics. The dataset contains $42$ different languages, of which $17$ are not present on existing multilingual article-level disinformation datasets presented in Table~\ref{tab:datasets}. Please refer to the supplementary material for a detailed breakdown of all languages.
Out of all $2,946$ publishers in the dataset, $1,187$ and $1,759$ are associated with disinformation and trustworthy articles, respectively. 
The top five publishers of disinformation articles are known pro-Kremlin outlets: Sputnik ( $15.7\%$), RT ($11.3\%$), RIA Novosti ($4.7\%$), Tsargrad ($2.1\%$), and Ukraina.ru ($1.6\%$). For trustworthy articles, the top five publishers include Reuters ($6.7\%$), BBC ($6.5\%$), The Guardian ($5.3\%$), Deutsche Welle ($3.9\%$), and Radio Free Europe ($3.2\%$).

\subsection{Analysis of Topics}
Figure \ref{fig:topics_by_language} 
shows the distribution of the eight most frequent topics across the whole dataset for the five most common languages. Notably, the topic of ``War in Ukraine'' is the most prevalent across all five languages. However, certain topics appear more frequently in specific languages.
The topic ``Invasion of Ukraine'' is over twice more common in Spanish than in other languages. ``Conspiracy theory'' and ``Crimea'' are frequently discussed in four out of the top five languages, except Spanish, where they account for only $2.1\%$ and $2.6\%$ of articles respectively, compared to an average of $11.6\%$ and $11\%$ in other languages. A similar discrepancy is noted for "Invasion of Ukraine" in German, appearing in just $1.1\%$ of articles, while averaging $12.5\%$ in Russian, Spanish, Arabic, and English. These disparities suggest a language-specific targeted dissemination of disinformation contingent on the topics.

\begin{figure}[ht]
\centering
  \begin{adjustbox}{width=0.93\columnwidth}
    \includegraphics{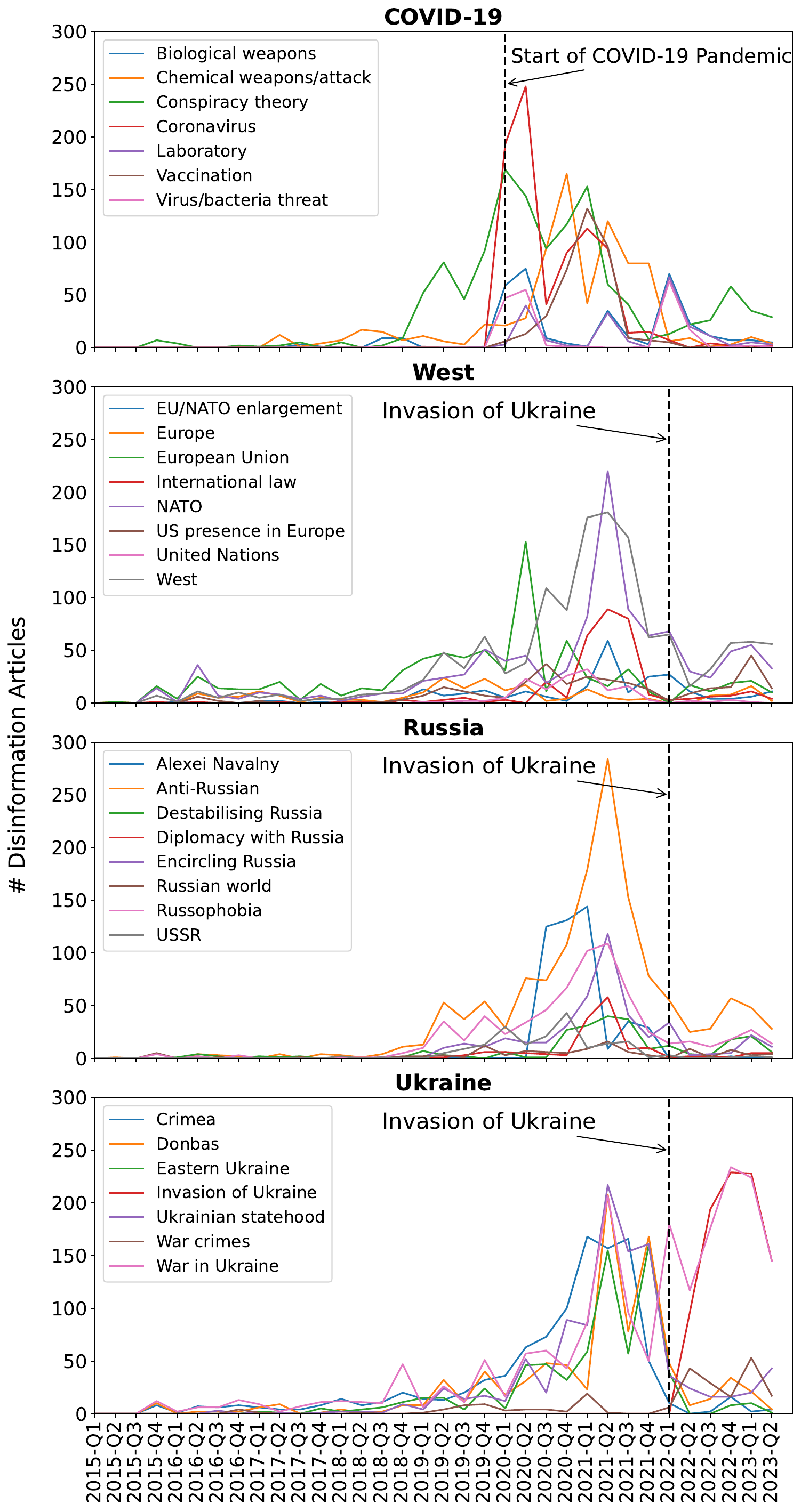}
  \end{adjustbox}
  \caption{Temporal distribution of disinformation topics.}
  \label{fig:topics_by_time}
\end{figure}

To examine the emergence of related themes over time, we group similar high-frequency topics into 
four overarching themes: ``COVID-19'', ``West'', ``Russia'', and ``Ukraine''. These themes are manually defined by merging similar topics from the top 50 most frequent in the dataset. Figure~\ref{fig:topics_by_time} shows the temporal distribution of disinformation topics under each theme. From the first quarter of 2021 onwards, there is a noticeable increase in disinformation articles across all themes, coinciding with Russia's escalating military presence along the Ukrainian border. Similar surge in the number of ``COVID-19''-related topics can be observed in the months following the start of the pandemic. After the full-scale invasion of Ukraine in the first quarter of 2022, disinformation articles on all topics decline significantly, except for "War in Ukraine" and "Invasion of Ukraine", which become the dominant disinformation topics.

\section{Experimental Setup} \label{sec:experimental_setup}
We employ the EUvsDisinfo dataset to perform binary disinformation detection (i.e. classify news articles as either disinformation or trustworthy). While the full dataset contains articles in $42$ unique languages, several languages are severely underrepresented and/or highly skewed towards one of the classes. Therefore in order to draw reliable conclusions, we discard underrepresented languages (fewer than $25$ articles) and those in which more than 95\% of the articles belong to only one of the two classes. The resulting dataset used in the experiments comprises $14,063$ articles distributed across $14$ languages (\% of disinformation articles is also shown)\footnoteref{sup}: English ($6,546$, $6\%$), Russian ($5,825$, $92\%$), German ($313$, $69\%$), French ($292$, $57\%$), Spanish ($287$, $85\%$), Georgian ($156$, $94\%$), Czech ($152$, $73\%$), Polish ($147$, $30\%$), Italian ($103$, $83\%$), Lithuanian ($78$, $36\%$), Romanian ($68$, $25\%$), Slovak ($35$, $91\%$), Serbian ($31$, $87\%$), Finnish ($30$, $27\%$).






We experiment with four models that have been used in prior work on article-level disinformation classification. We train commonly adopted baselines from previous work \cite{shu2020fakenewsnet, golbeck2018fake, yichuan-et-al-mmcovid, solopova-etal-2023-evolution}: Multinomial Naive Bayes (\texttt{MNB}), and Support Vector Machine (\texttt{SVM}), both using bag-of-words to encode the textual features. 
We also train transformer-based models, \texttt{mBERT} and \texttt{XLM-RoBERTa}, as three of the relevant multilingual datasets discussed in Section~\ref{sec:intro} use \texttt{mBERT} as a baseline \cite{momchil-et-al-2016-tallip, solopova-etal-2023-evolution, shahi2006multilingual}, and \texttt{XLM-R} is used by \citet{yichuan-et-al-mmcovid}. 
We conduct a hyperparameter search for each model by: (i) splitting the dataset into train and development sets with $90\%$ and $10\%$ of the data, respectively. The highest scoring hyperparameter configuration with respect to the development set is used throughout the experiments. (ii) discarding the development set, and splitting the remaining $90\%$ of the dataset into train and test sets using a 10-fold cross-validation strategy. The folds are stratified by both language and class. Model performance for each language is measured using the $F1\textnormal{-}Macro$ score to account for the skewed distribution of classes. To obtain a unified metric that encapsulates the overall system performance across all languages, we further average the $F1\textnormal{-}Macro$ scores per language to obtain the \Fmacroavg\ score. Further details such as hyperparameters and hardware configurations can be found in the supplementary material\footnoteref{sup}.

\section{Results} \label{sec:results}

\begin{table}[ht]
\caption{Classification results (Mean ± STD F1\textnormal{-}Macro). Best scores are in bold.}
\scalebox{0.95}{
\begin{tabular}{@{}ccccc@{}}
\toprule
\textbf{Language} & \textbf{\texttt{MNB}} & \textbf{\texttt{SVM}} & \textbf{\texttt{mBERT}} & \textbf{\texttt{XLM-R}} \\ \midrule
EN & $0.49$\scriptsize{±$0.01$} & $0.82$\scriptsize{±$0.03$} & \boldmath{$0.89$\scriptsize{±$0.03$}} & $0.86$\scriptsize{±$0.03$} \\
RU & $0.49$\scriptsize{±$0.02$} & $0.76$\scriptsize{±$0.05$} & \boldmath{$0.83$\scriptsize{±$0.04$}} & $0.82$\scriptsize{±$0.03$} \\
DE & $0.46$\scriptsize{±$0.11$} & $0.83$\scriptsize{±$0.08$} & \boldmath{$0.92$\scriptsize{±$0.05$}} & $0.86$\scriptsize{±$0.05$} \\
FR & $0.65$\scriptsize{±$0.13$} & \boldmath{$0.85$\scriptsize{±$0.07$}} & $0.83$\scriptsize{±$0.08$} & $0.83$\scriptsize{±$0.10$} \\
ES & $0.46$\scriptsize{±$0.03$} & $0.80$\scriptsize{±$0.15$} & \boldmath{$0.92$\scriptsize{±$0.11$}} & $0.85$\scriptsize{±$0.12$} \\
KA & $0.73$\scriptsize{±$0.27$} & \boldmath{$0.85$\scriptsize{±$0.26$}} & $0.73$\scriptsize{±$0.22$} & $0.66$\scriptsize{±$0.22$} \\
CZ & $0.42$\scriptsize{±$0.04$} & $0.82$\scriptsize{±$0.12$} & $0.83$\scriptsize{±$0.12$} & \boldmath{$0.88$\scriptsize{±$0.10$}} \\
PO & $0.71$\scriptsize{±$0.19$} & $0.78$\scriptsize{±$0.14$} & \boldmath{$0.88$\scriptsize{±$0.08$}} & $0.82$\scriptsize{±$0.12$} \\
IT & $0.56$\scriptsize{±$0.24$} & $0.80$\scriptsize{±$0.23$} & \boldmath{$0.81$\scriptsize{±$0.15$}} & $0.78$\scriptsize{±$0.15$} \\
LT & $0.65$\scriptsize{±$0.22$} & $0.72$\scriptsize{±$0.14$} & \boldmath{$0.89$\scriptsize{±$0.15$}} & $0.78$\scriptsize{±$0.18$} \\
RO & $0.57$\scriptsize{±$0.30$} & $0.52$\scriptsize{±$0.32$} & $0.73$\scriptsize{±$0.25$} & \boldmath{$0.88$\scriptsize{±$0.16$}} \\
SK & $0.83$\scriptsize{±$0.29$} & $0.83$\scriptsize{±$0.29$} & $0.86$\scriptsize{±$0.25$} & \boldmath{$0.92$\scriptsize{±$0.19$}} \\
SR & $0.76$\scriptsize{±$0.32$} & $0.76$\scriptsize{±$0.32$} & \boldmath{$0.82$\scriptsize{±$0.29$}} & $0.73$\scriptsize{±$0.29$} \\
FI & $0.71$\scriptsize{±$0.33$} & $0.63$\scriptsize{±$0.34$} & $0.66$\scriptsize{±$0.30$} & \boldmath{$0.77$\scriptsize{±$0.31$}} \\ \midrule
AVG & $0.61$ & $0.77$ & \boldmath{$0.83$} & $0.82$ \\ \bottomrule
\end{tabular}}
\label{tab:results_baselines}
\end{table}

Table~\ref{tab:results_baselines} shows the classification results for the proposed baselines.
\texttt{mBERT} achieves the highest average score ($0.83$), and the highest per-language score in $8$ out of the $14$ languages: English, Russian, German, Spanish, Polish, Italian, Lithuanian, and Serbian. Next,  \texttt{XLM-R} achieves an \Fmacroavg\ score of $0.82$, which is $1.2\%$ lower than that of \texttt{mBERT}, despite \texttt{XLM-R} having roughly $20$ times more parameters. The standard deviations for \texttt{mBERT} and \texttt{XLM-R} show that their scores largely overlap 
for most languages. Nonetheless, \texttt{XLM-R} achieves the highest scores for Czech, Romanian, Slovak, and Finnish. The \texttt{SVM} baseline achieves an \Fmacroavg\ score of $0.77$, a decrease of $7.2\%$ compared to \texttt{mBERT}. Surprisingly, the \texttt{SVM} baseline achieves the highest scores for two languages: French and Georgian. Lastly, the \texttt{MNB} model scores the lowest with an \Fmacroavg\ score of $0.61$, a significant $26.5\%$ decrease compared to \texttt{mBERT}.

\section{Conclusions and Future Work} \label{sec:conclusion}
This paper introduced EUvsDisinfo, a large, linguistically, temporally, and topically diverse dataset of 
disinformation and trustworthy articles originating from pro-Kremlin and reliable / less biased outlets, respectively.
Using the dataset, we found evidences of language-specific targeting of specific topics, and revealed a surge in disinformation content related to the war in Ukraine right before its full-scale invasion in 2022. Lastly, we proposed classification baselines using our dataset for the task of binary disinformation detection in a multilingual setting. In future work, we plan to leverage the structure of our dataset to explore evidence-aware fact-checking approaches by linking disinformation and trustworthy articles referring to the same narrative.

\section*{Ethical Statement}
The dataset is seeded from the publicly available debunks published by EUvsDisinfo. The authors of the manuscript are not part of the EUvsDisinfo organisation. The dataset content originates from news articles, thus it does not contain 
personal data. 
The dataset specifically targets pro-Kremlin disinformation; while pro-Western disinformation is not within the scope of this work.
We recognise that the dataset is susceptible to misuse for malicious purposes (e.g., used by originators of disinformation to improve their techniques), and we strongly urge researchers to use it in accordance with best practice ethics protocols. Our dataset is compliant with FAIR principles \cite{wilkinson2016fair}. It is made fully available with a unique digital object identifier, in a CSV format that can be processed by most widely used tools, and is released under an Apache 2.0 license. Since the textual content of news articles may be copyrighted, we do not include them in the dataset, but instead provide a software (\url{https://doi.org/10.5281/zenodo.10492913})
to allow researchers to collect the content themselves.



\begin{acks}
This  work  has  been  co-funded  by  the  UK’s innovation agency (Innovate UK) grant 10039055 (approved under the Horizon Europe Programme as vera.ai,  EU  grant  agreement  101070093) under action number 2020-EU-IA-0282. 
\end{acks}

\bibliographystyle{ACM-Reference-Format}
\balance
\bibliography{custom}

\end{document}